%% file: main.tex
\documentclass{article}
\usepackage{spconf,amsmath,graphicx}
\usepackage{amsfonts, amssymb, xcolor, dsfont}
\usepackage{url}

\usepackage{multirow}
\usepackage{subcaption}
\usepackage{booktabs,dcolumn,array}                 
\usepackage{colortbl}
\usepackage{graphicx}
\usepackage{tikz}
\usepackage{pgfplots}
\usepackage[mode=buildnew]{standalone}
\pgfplotsset{compat=1.18}

\DeclareMathOperator{\cut}{\operatorname{Cut}}
\DeclareMathOperator{\ncut}{\operatorname{NCut}}
\DeclareMathOperator{\vol}{\operatorname{Vol}}

\def\-{\scalebox{1}[1.0]{\( - \)}}

\title{A Sparse Graph Formulation for Efficient Spectral Image Segmentation }
\twoauthors
 {Rahul Palnitkar}
	{Haverford College \\ Department of Computer Science \\ 370 Lancaster Avenue \\ Haverford, PA 19041}
 {Jeova Farias Sales Rocha Neto}
	{Bowdoin College \\ Department of Computer Science \\
	255 Maine Street\\
	Brunswick, ME 04011}

\let\oldbibliography\thebibliography
\renewcommand{\thebibliography}[1]{%
  \oldbibliography{#1}%
  \setlength{\itemsep}{.6pt}%
  \setlength{\lineskiplimit}{-\maxdimen}
}

\begin{document}
%
\maketitle
\begin{abstract}
Spectral Clustering is one of the most traditional methods for solving segmentation problems. Based on Normalized Cuts, it partitions an image using an objective function defined by a graph. Despite their mathematical attractiveness, spectral approaches have traditionally been neglected by the scientific community because of their practical issues and underperformance.  In this paper, we adopt a sparse graph formulation based on the inclusion of extra nodes in a simple grid graph. While the grid encodes the spatial disposition of the pixels, the extra nodes account for the pixel color data. Applying the original Normalized Cuts algorithm to this graph leads to a simple and scalable method for spectral image segmentation, with a rich interpretable solution. Our experiments also show that our proposed methodology overperforms both traditional and modern unsupervised algorithms for segmentation in both real and synthetic data, establishing competitive results for many unsupervised segmentation datasets.
\end{abstract}
\section{Introduction}
\label{sec:intro}
Image Segmentation refers to the problem of grouping image pixels in meaningful regions with the goal of understanding or summarizing the visual content of that image \cite{szeliski2022computer}. Because of that, image segmentation has found applications in many domains, such as medical image analysis, autonomous driving, and photo editing, to name a few \cite{szeliski2022computer}. This intense industrial appeal led to the development of many algorithms that tackled the problem using variational \cite{grady2005multilabel}, graphical \cite{shi2000normalized, tang2013grabcut}, statistical \cite{rocha2022direct} and, more recently, deep learning techniques \cite{minaee2021image}.

Normalized Cuts ($\ncut$) \cite{shi2000normalized} provides one of the most classical algorithms for image segmentation, and it stands out for its methodological consistency backed by graph theory and computational linear algebra. It consists of encoding the image data into a graph and then using spectral techniques to approximately find the cut corresponding to the final segmentation. It has been a lasting algorithm for vision problems \cite{fowlkes2004spectral} and is currently receiving prodigious research interest due to the emergence of Vision Transformers \cite{han2022survey}, due to its ability to automatically detect objects and segment images using Transformer-based image embeddings, such as those arising from self-distillation (DINO) approaches \cite{oquab2023dinov2, melas2022deep, wang2022tokencut}. 

Despite its persistent theoretical appeal, $\ncut$'s initial formulation rapidly met with drawbacks that would hinder its broader adoption by the scientific community, although many strategies were approached to overcome them \cite{fowlkes2004spectral, wu2018scalable}. Historically, the issues were mainly (1) computational, as it required handling large matrices \cite{fowlkes2004spectral, chew2015semi}, and (2) practical, as it performed worse than other unsupervised segmentation algorithms \cite{arbelaez2010contour, maji2011biased}. Some of these issues were amended with the use of rich and low-resolution self-supervised transformer features \cite{caron2021emerging} at the pixel level, followed by the standard $\ncut$ approach \cite{melas2021finding, wang2022tokencut}. Despite that, these methods still scale poorly, only being shown to perform well on the low-resolution scale typical of features arising from vision transformers, often needing post-processing algorithms such as Conditional Random Fields (CRF) or Bilateral Solvers \cite{wang2022tokencut} for edge refinement. Finally, none of these approaches considers the interpretability of their optimization problem.


In this paper, we address all these issues by adopting a sparse graph consisting of a grid graph that encodes the pixel spatial locations and a  few extra nodes corresponding to the pixel intensities \cite{grady2005multilabel}. We show that the $\ncut$ solution found  using this graph is \textit{interpretable}, as it encourages pixel groupings that linearly balance spatial coherence 
(neighbors generally belong to the same group) and color cohesion (pixels in a segment have colors in 
the same subset of colors) \cite{rocha2020spectral}. We also demonstrate that this simple spectral formulation is \textit{scalable}, and that \textit{its only parameter has a clear interpretation}, which facilitates its tuning. Finally, we show that using our methodology to segment images based on their extracted DINO features outperforms recent approaches to use this data in the context of spectral segmentation \cite{kim2023causal, wang2022tokencut}. We compare our method with other traditional and modern unsupervised techniques for segmentation and observe that \textit{our simple methodology outperforms its spectral counterparts} in terms of runtime and quality, while also being competitive against the other methods. 


    \setlength{\belowdisplayskip}{4pt} \setlength{\belowdisplayshortskip}{4pt}
    \setlength{\abovedisplayskip}{4pt} \setlength{\abovedisplayshortskip}{4pt}

\section{Background}
\label{sec:format}

\subsection{Normalized Cuts}
Let $I$ be an image of $n$ pixels and $k$ unique colors. Let $G = (V, E, w)$ be an undirected weighted graph where each node corresponds to a pixel, and the weight function $w(i,j)$ evaluates the similarity between pixels $i$ and $j$. A binary segmentation of $I$ is achieved by finding a partition $(A, B)$ of $V$ that minimizes the cut value $\cut (A, B) = \sum_{i \in A, j \in B} w(i, j)$. 

This approach however favors unbalanced partitions, leading to undesirable solutions. A way of addressing this is to consider the normalized cut value $\ncut (A, B)$ \cite{shi2000normalized} instead,
\begin{equation*}
    \ncut (A, B) = \frac{\cut(A, B)}{\vol(A)} + \frac{\cut(A, B)}{\vol(B)},
\end{equation*}
where $\vol(A) = \sum_{i \in A, j \in V} w(i, j)$ is the ``volume'' of $A$. We use $\vol(V) = \vol(A) + \vol(B)$ to represent the volume of $G$. Let $W$ be the weighted adjacency matrix of $G$ and $D$ be the diagonal degree matrix with $D(i,i) = \sum_{j \in V} w(i,j)$.  The \emph{Laplacian} matrix of $G$ is then defined as $L=D-W$. The problem of minimizing the $\ncut$ of $G$ becomes:
\begin{equation}\label{eq:ncutunrelaxed}
\begin{aligned}
\min_{\mathbf{x}} \quad & \mathcal{E}(\mathbf{x}|G) ,\\
\textrm{s.t.} \quad & {\mathbf{x}^{\top}D\mathbf{x}} = 1, \quad \mathbf{1}^{\top}D\mathbf{x} = 0, \quad \mathbf{x} \in \left\{-\alpha, \beta\right\}^n.   \\
\end{aligned}
\end{equation}
\noindent where $\mathbf{x}$ is the vector that corresponds to our final segment assignment, $\mathcal{E}(\mathbf{x}|G) = \mathbf{x}^{\top}L\mathbf{x}$ is the objective function or \textit{assignment energy} we wish to minimize and:
\begin{equation}\label{alphabeta}
    \alpha = \sqrt{\frac{\vol(A)}{\vol(V)\vol(B)}}, \quad \beta = \sqrt{\frac{\vol(B)}{\vol(V)\vol(A)}}.
\end{equation}
The algorithm in \cite{shi2000normalized}, here called the $\ncut$ algorithm, solves a relaxation of the minimum $\ncut$ problem by dropping the integer constraint in Eq. \ref{eq:ncutunrelaxed}. The new optimization problem can be solved via generalized eigenvector decomposition,
\begin{equation}\label{eqn:geigenvector}
L\mathbf{x} = \lambda D\mathbf{x}.
\end{equation}
The algorithm then selects the eigenvector $\mathbf{x}$ with second-smallest
eigenvalue, and partitions $V$ by thresholding $\mathbf{x}$ \cite{shi2000normalized}. 

\subsection{Traditional and Recent Approaches for Edge Weighting and their Drawbacks}
A crucial step in developing efficient segmentation solvers using $\ncut$ is to define how the function $w$ is computed. In the original $\ncut$ work  and in early subsequent work \cite{zelnik2004self, tao2007color}, the $w$ combined two grouping cues in a single value:
\begin{equation}\label{eq:tradweight}
\begin{aligned}
 w(i,j)=& \exp\left(-\frac{\lVert I(i) \- I(j)\rVert ^2}{2\sigma_I^2}-\frac{\lVert X(i) \- X(j)\rVert ^2}{2\sigma_X^2}\right)   \\ &\times 
 \mathds{1}(\lVert X(i) - X(j)\rVert < r ),
\end{aligned}
\end{equation}

\noindent where $I(j)$ and $X(j)$ are the appearance (such as the color or other pixel-level semantic feature) of pixel $j$ and its spatial location on image $I$, respectively, $\mathds{1}(\cdot)$ is the indicator function, and $r$, $\sigma_I$ and $\sigma_X$ have fixed values.

The $\ncut$ formulation based on the above weight function presents several drawbacks. For one, it is computationally inefficient to construct the graph in Eq. \ref{eq:tradweight} for high-resolution images, since it requires visiting each pixel and their neighboring locations within a radius $r$. Secondly, the graph is potentially dense, which poses memory and runtime issues during eigenvector computation. $\ncut$ with weights as in Eq. \ref{eq:tradweight} is also known to be underperforming in practice \cite{arbelaez2010contour} and calibrating the values for $\sigma_X$ and $\sigma_I$ is challenging \cite{rocha2020spectral}. That led to the development of spectral image segmentation methods that (1) use superpixels \cite{achanta2012slic} instead of working on the original pixel set for algorithmic efficiency and (2) employ contour-based information, such as Probability of Boundary (Pb) \cite{arbelaez2010contour} instead of raw color intensities \cite{maji2011biased, chew2015semi} for performance. Finally, the optimization in Eq. \ref{eq:ncutunrelaxed} on the graph constructed using Eq. \ref{eq:tradweight} lacks interpretability \cite{rocha2020spectral}, i.e., all the understanding one can conceive out of its solutions is that it approximately finds segments whose pixels are similar in appearance and especially close enough to each other. Ideally, one would also be concerned about the extent to which certain parameters and each constituent data component (such as the pixel appearance data and its location information) affects the algorithm's performance.

More recently, researchers took advantage of the performant but low-resolution features from self-supervised DINO architectures \cite{caron2021emerging}, to derive fully connected graphs whose weights are mostly given by simple cosine similarities \cite{melas2022deep, wang2022tokencut}. Although the resulting eigenvectors can be used to achieve good results after rescaling, such methodologies still rely on their low-resolution setting to be computationally efficient. Furthermore, this approach does not address the lack of interpretability discussed above.

\begin{table*}
    \centering
    \caption{Results on $100\times 100$ synthetic images arising from different binary patterns. We use RGB features in each method. We outperform in runtime and mIoU classical spectral segmentation techniques in many challenging noise and pattern settings.}
    \vspace{-5pt}
    \includegraphics[width=\textwidth]{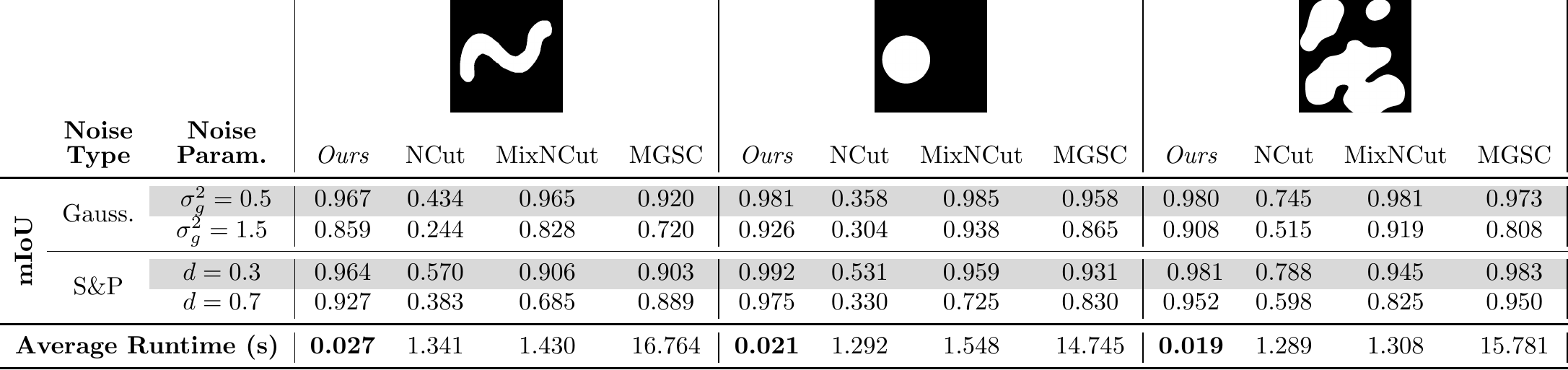}   
     \vspace{-15pt}
    \label{fig:quant_synt}
\end{table*}

\section{Proposed Methodology}
\label{sec:pagestyle}

\subsection{New Graph Formulation}
Inspired by \cite{grady2005multilabel} and \cite{tang2013grabcut}, we show that a sparse graph formulation efficiently addresses the issues above. Our graph construction starts by considering a grid graph $G_{\text{grid}}$ whose vertices correspond to image pixels, with 
 edges of weight $\mu > 0$ between each pair of neighboring pixels. 
 From $G_{\text{grid}}$, we create a new graph $G_\text{all}$ by adding $k$ extra nodes, each corresponding to one of the $k$ unique colors in $I$\footnote{What is described here can be generalized to consider $k$ clusters of pixel-level features, instead of $k$ unique colors. This will be important when handling images where the number of unique colors is large or with multidimensional continuous pixel features, such as the ones arising from self-supervised vision transformers \cite{caron2021emerging}.}. Each node on $G_{\text{grid}}$ is then connected to the extra node corresponding to its color with an edge of unitary weight. This graph formulation is computationally attractive for its sparsity. Assuming $k \ll n$, $G_\text{all}$ only requires $O(n)$ edges. When segmenting $I$, we only make use of the assignment vector for the grid nodes.
\begin{figure}[t!]
    \input{imgs/synt/res_synt.tex}
    \vspace{-5pt}
    \caption{{Sample qualitative results on synthetic images.} Top: Gaussian noise, $\sigma_g^2 = 1.5$. Bottom: Salt \& Pepper, $d = 0.7$. 
     }
     \vspace{-5pt}
    \label{fig:quali_synt}
\end{figure}

\subsection{Interpretation of the Cut on $G_\text{all}$}

 Let  $W_{\text{grid}}$ and $L_{\text{grid}}$ be the adjacency and  Laplacian matrices of $G_{\text{grid}}$, respectively. Let $W_{\text{all}}$ and $L_{\text{all}}$ be the analogous to $G_\text{all}$. These matrices are related to each other as follows:
\begin{equation*}
W_{\text{all}} = 
\begin{bmatrix}
W_{\text{grid}} &  H \\
H^\top & 0 
\end{bmatrix},
\quad L_{\text{all}} = 
\begin{bmatrix}
L_{\text{grid}} +  I & - H \\
- H^\top &  D_{n} 
\end{bmatrix}
\end{equation*}
where $H \in \mathbb{R}^{n \times k}$ is a one-hot encoding of the colors in $I$, i.e. $H(i, j) = \mathds{1}\{I(i) = j\}$, and $D_n \in \mathbb{R}^{k \times k}$ is diagonal such that $D_n(j, j) = n_j$, where $n_j$ is the number of pixels in $I$ with color $j$. Let $\mathbf{x}$ and $\mathbf{y}$ be the assignment vectors for the nodes in $G_{\text{grid}}$ and for the extras nodes, respectively. Note that $\mathbf{x}\in \{-\alpha, \beta\}^n$ and $\mathbf{y} \in \{-\alpha, \beta\}^k$, where $\alpha$ and, $\beta$ are computed using Eq. \ref{alphabeta} on the graph $G_\text{all}$. Let $\mathbf{z} = [\mathbf{x}|\mathbf{y}]$ be $\mathbf{x}$ and $\mathbf{y}$ concatenated. The objective in Eq. \ref{eq:ncutunrelaxed} then becomes:
\begin{equation}
    \begin{aligned}
    \mathcal{E}(\mathbf{z}|G_\text{all}) &= \mathbf{z}^{\top}L_\text{all}\mathbf{z} \\&= \mathbf{x}^\top (L_{\text{grid}} +  I)\mathbf{x} - 2 (H\mathbf{y})^\top\mathbf{x} + \mathbf{y}^\top D_{n}\mathbf{y}.\label{eq:Ez}
\end{aligned}
\end{equation}
Noticing that $D_n = H^\top H$, we derive the energy associated with the assignment $\mathbf{z}$ of the nodes on $G_\text{all}$:
\begin{equation}\label{eq:energy_all}
    \mathcal{E}(\mathbf{z}|G_\text{all}) = \lVert \mathbf{x} - H \mathbf{y}\rVert^2_2 + \mathbf{x}^{\top}L_{\text{grid}}\mathbf{x}.
\end{equation}
The first term in the above expression accounts for the number of pixels that are assigned to a partition that is not the same as the partition its color is assigned to. In fact, if we define $n_{\text{missmatch}}$ as the number of these mismatches, we have:
\begin{equation}
    \lVert \mathbf{x} - H \mathbf{y}\rVert^2_2 = (\alpha+\beta)^2 n_{\text{missmatch}}.
\end{equation}
\noindent Because of the grid nature of $G_{\text{grid}}$, the second term in Eq. \ref{eq:energy_all} accounts for the number of neighboring pixel pairs that are assigned to different segments:
\begin{equation}
    \mathbf{x}^{\top}L_{\text{grid}}\mathbf{x} = \mu (\alpha + \beta)^2n_{\text{boundary}},
\end{equation}
\noindent where $n_{\text{boundary}}$ stands for the number of pixels on the boundary of the final segmentation. Finally, noticing that $ (\alpha+\beta)^2=\vol(V)/(\vol(A)\vol(B))$ from Eq. \ref{alphabeta}, we have that:
\begin{equation}\label{eq:energy_all2}
    \mathcal{E}(\mathbf{z}|G_\text{all}) = \frac{\vol(V)}{\vol(A)\vol(B)}(n_{\text{missmatch}} + \mu n_{\text{boundary}}).
\end{equation}
\noindent Minimizing $\mathcal{E}(\mathbf{z}|G_\text{all})$ for $\mathbf{x}$ and $\mathbf{y}$, results in finding a partition of $G_\text{all}$ 
that trades off between spatial coherence (neighboring pixels should belong to the same segment) and color cohesion (pixels in a segment should have colors belonging to the same subset of available intensities) via the value of $\mu$. Minimizing the multiplicative factor in Eq. \ref{eq:energy_all2} also encourages a balanced partition of the nodes in $G_{\text{all}}$.

\begin{figure}[b]
    \vspace{-5pt}
    \input{data/vary_size_code}
    \caption{Segmentation results for varying image sizes. 
    Left: a sample image for $\ell = 500$. 
    Right: runtime and mIoU results varying the number of image pixels $n = \ell^2$.} 
    \label{fig:vary_size}
\end{figure}

\begin{table*}[h!]
    \centering
    \caption{Average mIoU on BSDS500. We provide results for our method on RGB intensities or DINO pixel-level features. Our proposed method is shown to overperform many state-of-the-art unsupervised segmentation algorithms even in its simple RGB-based formulation. \dag The method listed was re-trained using code made available by their authors.}
    \vspace{-5pt}
    \setlength\tabcolsep{5pt} 

    \begin{tabular}{cccccccccccc}
        \toprule
            $\ncut$\dag & $K$-means & GS & IIC  &  DFC\dag  & DoubleDIP\dag   & PiCIE\dag   & W-Net\dag & SegSort\dag  & \textit{Ours} (RGB) & \textit{Ours} (DINO) \\\midrule
          0.385  & 0.240 & 0.313 & 0.172  & 0.305  & 0.356   & 0.325    & 0.428 & 0.480 & 0.479 & 0.534  \\\bottomrule
    \end{tabular}
    \label{tab:quantitative_comparison_real_BSDS500}
\end{table*}


Finally, one can spot some formulaic similarities between the expression in Eq. \ref{eq:energy_all} and typical optimization objectives in inverse problem solvers \cite{benning2018modern}, such as those involving the compromise of a quadratic data fidelity term and a regularization functional. Despite that similarity not being suitable for a complete interpretation of  Eq. \ref{eq:energy_all}, since here we have both $\mathbf{x}$ and $\mathbf{y}$ as unknowns, it can still aid our understanding of the minimization carried out in our proposed graph. Minimizing Eq. \ref{eq:energy_all} also corresponds to finding an assignment vector $\mathbf{z}$ where, on one hand, the extra node cluster assignments $\mathbf{y}$ are responsible as much as possible to the labeling of the pixel nodes $\mathbf{x}$ (first term), while also encouraging neighboring pixel labels to agree among themselves (second term).



\section{Numerical Experiments}\label{sec:exp}

\subsection{Experimental Setup}

We proceed with experiments on real and synthetic images. For our synthetic results, we contaminate three different background and foreground square patterns (intensities of either 0 and 1) of side length $\ell$ with either additive zero mean and variance $\sigma_g^2$ Gaussian noise or Salt \& Pepper (S\&P) with noise density $d$. The generated images are then rescaled and discretized, so the color intensities are integers in $[0, 255]$. We also test our method using all images in BSDS500 \cite{arbelaez2010contour}, DUT-OMRON \cite{yang2013saliency}, DUTS \cite{wang2017} and ECSSD \cite{shi2015hierarchical} segmentation datasets. We test segmenting them solely based on either their RGB data or their DINOv2 \cite{oquab2023dinov2} features interpolated to the pixel level. In practice, we perform PCA on the extracted DINO features and the first 3 principal dimensions to be interpolated. The choice of 3 components is inspired by the original work in \cite{oquab2023dinov2}. For each image, we vectorize their pixel feature space using mini-batch $K$-means clustering \cite{cho2021picie} with $k = 256$ as the only preprocessing stage for our proposed method. Unless otherwise specified, we set $\mu = 0.01$. In our methods, we do not apply any edge refinement post-processing.

Since the images in our real dataset may contain multiple objects/segments in their ground-truth, we generalized our proposed binary segmentation method taking the prescription in \cite{shi2000normalized} as inspiration. Here, we recursively reapply our proposed method on the partitions found by it until the energy $\mathcal{E}(\mathbf{z}|G_\text{all})$ (ref. Eq. \ref{eq:Ez}) of that partition is below given a threshold (set to 0.01) or until a given partition had already been subdivided 3 times.

\begin{table}[b]
\centering
\caption{Average mIoU on binary segmentation datasets.}\label{tab:binary_seg}
\begin{tabular}{ccccc}
\toprule
Method  & \hspace*{1.5mm} DUTS \hspace*{1.5mm} & \hspace*{1.5mm} ECSSD \hspace*{1.5mm} & DUT-OMRON \\\midrule
LOST & 0.518 & 0.654 & 0.410 \\
DeepUSPS & 0.305 & 0.440 & 0.305 \\
BigBiGAN & 0.498 & 0.672 &  0.453 \\
DeepSpectral & 0.514 & 0.733 & 0.567 \\
TokenCut & 0.576 &  0.712  & 0.533 \\
\textit{Ours} (RGB) & 0.577 &  0.617  & 0.509 \\
\textit{Ours} (DINO) & 0.569 &  0.684  & 0.588 \\\bottomrule
\end{tabular}
\end{table}

\begin{figure*}
    \input{imgs/vary_lambda/vary_lambda.tex}
    \vspace{-5pt}
    \caption{Qualitative results on BSDS500 images for various $\mu$ using pixel RGB features. The proposed methods is performs fast image segmentation in various imaging settings despite its spectral nature. Its sole parameter ($\mu$) regulates the level of border detail in the final segmentation. (b)-(d) show the eigenvector found and (left) and the resp. segmentation (right). Both images are of size $481\times 321$ and the average runtime is shown in parentheses.}
    \label{fig:vary_lambda}
\end{figure*}


For performance evaluation, we follow the literature \cite{melas2022deep, wang2022tokencut, kim2023causal} and assess the segmentations via the intersection over union  $\text{IoU} = |R \cap S|/|R \cup S|$, where $R$ and $S$ are a ground truth image region and its predicted segmentation, respectively. We compute the mean IoU (mIoU) by averaging the IoU values over the image \cite{kim2020unsupervised}. The predicted segments are matched one-to-one with target segments using a version of the Hungarian algorithm modified to accommodate when the number of segments in the final segmentation is different from that of the ground truth. BSDS provides multiple ground-truth segmentations, so we average their respective mIoUs.

We compare our proposed method with $\ncut$ and two more spectral methods for unsupervised segmentation, Multi Graph Spectral Clustering (MGSC) \cite{zhou2007spectral} and MixNCut \cite{rocha2020spectral} in our synthetic experiments. For our real data, we compare our results to those of three traditional baselines, $K$-means, Graph-based Segmentation (GS) \cite{felzenszwalb2004efficient} and SLIC \cite{achanta2012slic}; and various deep learning-based unsupervised segmentation algorithms: W-Net \cite{xia2017w}, IIC \cite{ji2019invariant},  SegSort \cite{hwang2019segsort}, DeepUSPS \cite{nguyen2019deepusps}, DFC \cite{kim2020unsupervised}, LOST \cite{simeoni2021localizing}, PiCIE \cite{cho2021picie}, BigBiGAN \cite{voynov2021object}, DeepSpectral \cite{melas2022deep}, TokenCut \cite{wang2022tokencut} and ACSeg \cite{li2023acseg}. When not otherwise specified as re-trained, we used the quantitative results provided in \cite{kim2020unsupervised}, \cite{kim2023causal} and \cite{wang2022tokencut}. W-Net, DeepSpectral and TokenCut are of particular interest to us, as they use Normalized Cuts concepts to devise their segmentation techniques.

Parameter tuning and model selection are performed in each re-trained method. We refrain from any sparsification algorithm in any spectral method. For $\ncut$, we also set $r = \infty$ in Eq. \ref{eq:tradweight} for simplicity. In our synthetic experiments, we apply $\ncut$ on the full pixel set. In the experiments using RGB images, we first use SLIC \cite{achanta2012slic} to over-segment the image in 500 superpixels. We also apply the recursive strategy above for $\ncut$ when segmenting multi-region images.

The experiments ran on an NVIDIA GeForce RTX 4070 GPU with 8 Gb VRAM memory and were implemented\footnote{A demo code for this paper is available at \url{www.github.com/rahul-palnitkar/Sparse-Graph-Spectral-Segmentation}.} in Python 3.10 using the GPU-accelerated sparse linear algebra utils of CuPy 12.3.

\subsection{Results and Discussion}

In Table \ref{fig:quant_synt}, we show the quantitative assessment of our proposed method for images with $\ell = 100$ pixels and the two different noise contaminations when generating each synthetic image under the three proposed ground-truth patterns. We also display the average runtime for generating each of the best segmentations in each method per pattern. In Figure \ref{fig:quali_synt}, we qualitatively compare the best results of each method under two challenging noise settings. These results demonstrate that our method greatly outperforms its related spectral methods in terms of runtime, a consequence of its sparse nature, while also being competitive in terms of mIoU, attaining almost perfect segmentations in most challenging scenarios.


In Figure \ref{fig:vary_size}, we present our average performance for images arising from the three patterns in Figure \ref{fig:quant_synt}. We vary $\ell$ and contaminate each image with Gaussian noise, $\sigma_g^2 = \ell/100$, so they become increasingly noisier with size. These results demonstrate that our method scales well despite its spectral nature and generates high-quality segmentations on large images, even in challenging noise settings. The runtime growth depicted in Figure \ref{fig:vary_size} reflects the linear complexity of our method. 

In Figure \ref{fig:vary_lambda}, we evaluate the effect of $\mu$ on binary segmentation and qualitatively demonstrate our method's performance and runtime on two sample images using their RGB intensities. Here, we note that $\mu$ has the effective role of "blurring" the eigenvector and consequently smoothing the borders of the final segmentation, removing artifacts or boundary details as it increases. This clear and straightforward parameter interpretation is not present in the traditional formulations of the Normalized Cut problem \cite{melas2022deep}, which entails that tuning it for a specific need is also simpler. Finally, we note our method's low runtime in relatively large images without the need for any image approximation or interpolation despite its spectral nature.    

\begin{figure*}
    \centering
    \input{imgs/real/comp_real}
    \vspace{-5pt}
    \caption{Qualitative comparison of our method to other deep learning-based unsupervised segmentation methods on challenging BSDS500 images. Our proposed methodology is generally able to preserve borders well and, when using DINO features, qualitatively outperform other deep-learning based unsupervised segmentation techniques.}
    \label{fig:qualitative_comparison}
\end{figure*}

In Table \ref{tab:quantitative_comparison_real_BSDS500}, we quantitatively demonstrate the mIoU performance of our proposed method on BSDS images, comparing it to other unsupervised segmentation algorithms. We display results using RGB pixel intensities or DINO features interpolated to the pixel level. 
$\ncut$ and W-Net perform surprisingly well compared to the other baselines and outperform many Deep Learning methods on BSDS. This can be seen as evidence of the power of spectral formulations to unsupervised segmentation tasks. Our method, when applied to RGB data, is only outperformed, slightly, by SegSort, which despite being unsupervised still makes use of pre-trained deep learning features. This means that our simple methodology when using just color data performs as well as other methods that rely on richer, but harder-to-train, features, demonstrating the suitability and appeal of our formulation to segmentation. On the other hand, when also making use of rich semantic features such as those from DINO, our method outperforms all depicted methods on BSDS. 


 Table \ref{tab:binary_seg} considers our segmentation methodology in the single object segmentation setting. Through this table, we note the overwhelming benefit of using spectral methods in the context of DINO features compared to other deep learning solutions. Furthermore, we again demonstrate the superior performance of our method on RGB data, achieving better performance results than many deep-learning solutions backed by rich semantic features. Indeed, it attains a state-of-the-art unsupervised segmentation performance on DUTS. Using DINO features on our algorithm usually improves its performance and also makes it achieve superior results on DUT-OMRON compared to DeepSpectral and TokenCut, our spectral DINO-backed comparison algorithms. We, however, fail to outperform these same methods on the ECSSD dataset, and future research will consider the reasons for this underperformance and approach different algorithmic improvements in our method to improve it.

 Finally, Figure \ref{fig:qualitative_comparison} qualitatively compares our method in its RGB and DINO formats against a few deep-learning-backed algorithms on challenging images from the BSDS dataset. Our method, especially when using DINO data, can detect low-contrast objects without losing border details. Again, this contrasts with the classical NCut results from \cite{shi2000normalized}, where much of segmentation borders are smoothed out due to their underlying graphical choice. Using extranodes, we qualitatively show that our method preserves borders well, while also having lower runtime and memory complexities.

\section{Conclusion}
In this paper, we introduce a new spectral algorithm for image segmentation that overcomes some historical issues found in the $\ncut$ framework and its related methods. We employ an efficient sparse graph, whose usage under the traditional $\ncut$ optimization problem leads to an interpretable cut value and an efficient solver. We also show that our method, which is fully characterized by just one parameter, is fast and produces high-quality results under high-noise synthetic settings, and outperforms many state-of-the-art unsupervised image segmentation algorithms on popular image datasets, especially when backed by rich pixel-level features. These results demonstrate the effectiveness of using our sparse formulation for spectral image segmentation and provide further evidence for the successful use of spectral methods in self-supervised transformer data.

\bibliographystyle{IEEEbib}
\bibliography{main}

\end{document}

%% file: imgs/synt/res_synt.tex
 \begin{subfigure}{0.09\textwidth}
        \centering
        \includegraphics[width=\textwidth]{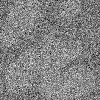}
        \\[2pt]
        \includegraphics[width=\textwidth]{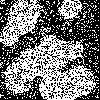}
        \caption*{Image}
    \end{subfigure}%
    \hfill
    \begin{subfigure}{0.09\textwidth}
        \centering
        \includegraphics[width=\textwidth]{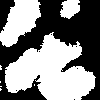}
        \\[2pt]
        \includegraphics[width=\textwidth]{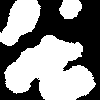}
        \caption*{\textit{Ours} (RGB)}
    \end{subfigure}%
    \hfill
    \begin{subfigure}{0.09\textwidth}
        \centering
        \includegraphics[width=\textwidth]{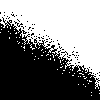}
        \\[2pt]
        \includegraphics[width=\textwidth]{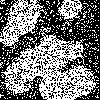}
        \caption*{NCut}
    \end{subfigure}%
    \hfill
    \begin{subfigure}{0.09\textwidth}
        \centering
        \includegraphics[width=\textwidth]{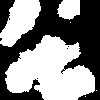}
        \\[2pt]
        \includegraphics[width=\textwidth]{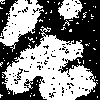}
        \caption*{MixNCut}
    \end{subfigure}%
        \hfill
    \begin{subfigure}{0.09\textwidth}
        \centering
        \includegraphics[width=\textwidth]{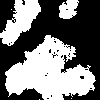}
        \\[2pt]
        \includegraphics[width=\textwidth]{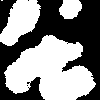}
        \caption*{MGSC}
    \end{subfigure}%

%% file: data/vary_size_code.tex
\vspace{-15pt}
\begin{subfigure}[t]{0.115\textwidth}
    \includegraphics[width=\textwidth]{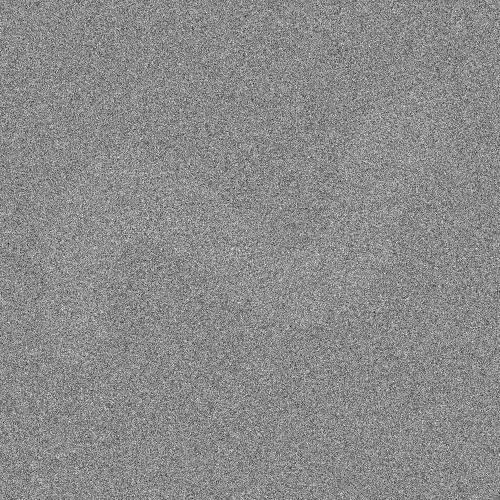}
\end{subfigure}%
\hspace{10pt}
\begin{subfigure}[t]{.3\textwidth}
    
    \begin{tikzpicture}
        \begin{axis}[
            scale only axis,
            height=1.25cm,
            width=.75\linewidth,
            ytick pos=left,
            xlabel={$n$},
            ylabel={mIoU},
            yticklabel style = {font=\footnotesize, color=red},
            xticklabel style = {font=\footnotesize},
            every y tick/.style={red},
            ylabel near ticks,
            xlabel near ticks,
            x label style={at={(axis description cs:0.5,-0.3)},anchor=north},
            xmin=10000,xmax=640000,
            ymin=0.6, ymax=1,
            scaled y ticks=true,
            yticklabel style={/pgf/number format/fixed},
            every x tick label/.append style={alias=XTick,inner xsep=0pt},
            every x tick scale label/.style={at=(XTick.base east),anchor=base west},
            scaled x ticks=false,
            scaled x ticks=manual:{\hspace{2pt} millions}{\pgfmathparse{#1/100000}},
            ]
       
            \addplot[mark=none, red] table[x index=1,y index=4] {data/size_times_average_variable2.dat};
        \end{axis}
        
        \begin{axis}[
            scale only axis,
            height=1.25cm,
            width=.75\linewidth,
            axis y line*=right,
            axis x line=none,   
            ylabel={time (s)},
            ymin=0.1, ymax=3.7,
            yticklabel style = {font=\footnotesize, color=blue},
            xticklabel style = {font=\footnotesize},
            every y tick/.style={blue},
            ylabel near ticks,
            xlabel near ticks,
            xmin=10000,xmax=640000,  
            y label style={at={(axis description cs:1.1,.6)}},
            ]
            \addplot[mark=none, blue] table[x index=1,y index=3] {data/size_times_average_variable2.dat};

        \end{axis}
    \end{tikzpicture}
\end{subfigure}

%% file: imgs/vary_lambda/vary_lambda.tex
   \begin{subfigure}[t]{0.13\textwidth}
        \includegraphics[width=\linewidth]{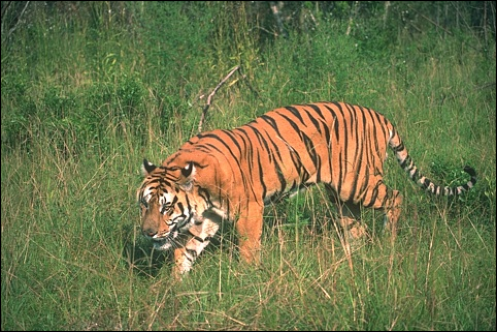}\hfill%
        \\[2pt]
        \includegraphics[width=\linewidth]{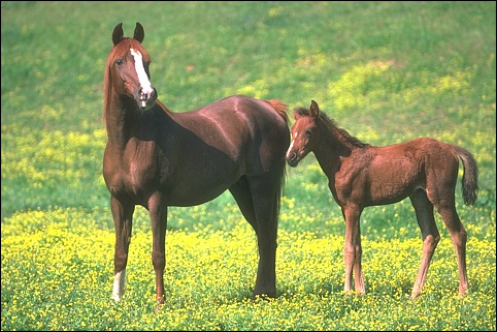}\hfill%
        \\[2pt]
        \includegraphics[width=\linewidth]{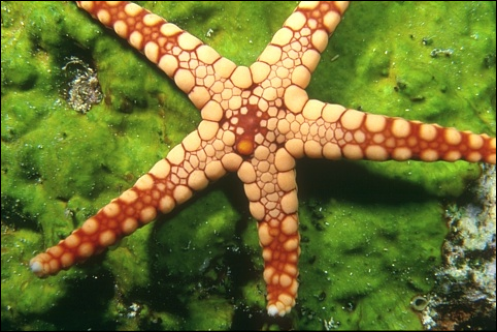}
        \caption{Image}
    \end{subfigure}%
    \hfill
    \begin{subfigure}[t]{0.27\textwidth}
        \includegraphics[width=.49\linewidth]{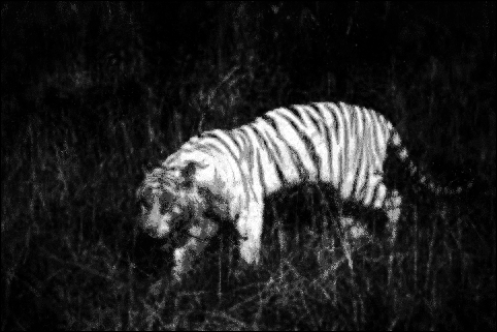}\hfill%
        \includegraphics[width=.49\linewidth]{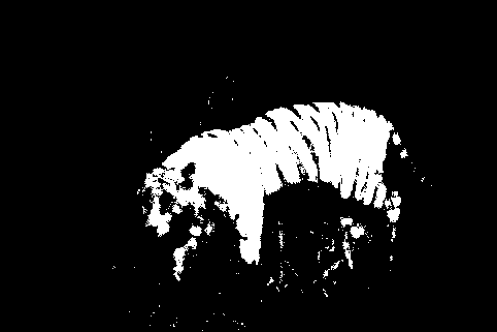}
        \\[1pt]
        \includegraphics[width=.49\linewidth]{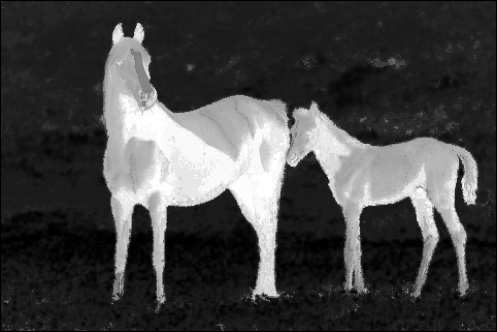}\hfill%
        \includegraphics[width=.49\linewidth]{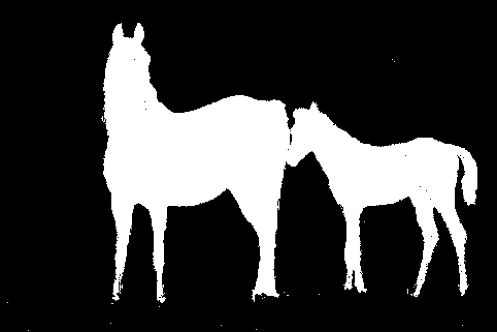}
        \\[1pt]
        \includegraphics[width=.49\linewidth]{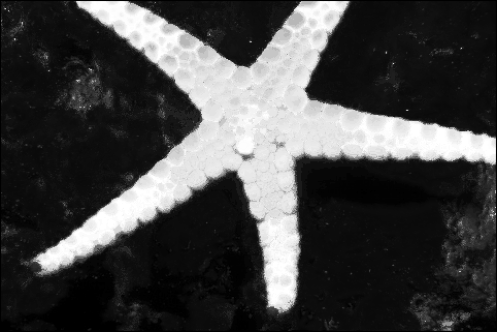}\hfill%
        \includegraphics[width=.49\linewidth]{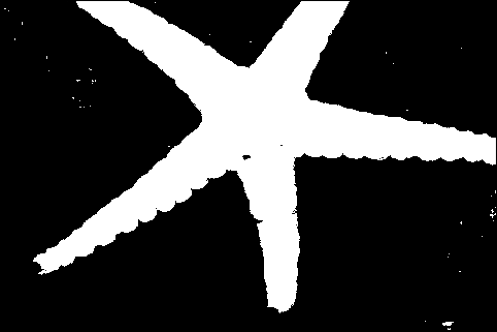}
        \caption{$\mu = 0.01$ ($0.30s$)}
    \end{subfigure}%
    \hspace{6pt}
    \begin{subfigure}[t]{0.27\textwidth}
        \includegraphics[width=.49\linewidth]{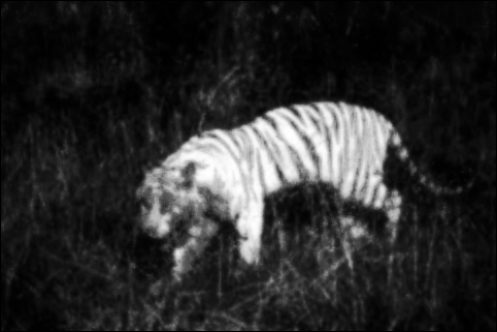}\hfill%
        \includegraphics[width=.49\linewidth]{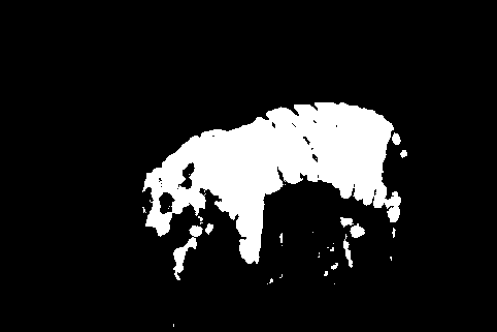}
        \\[1pt]
        \includegraphics[width=.49\linewidth]{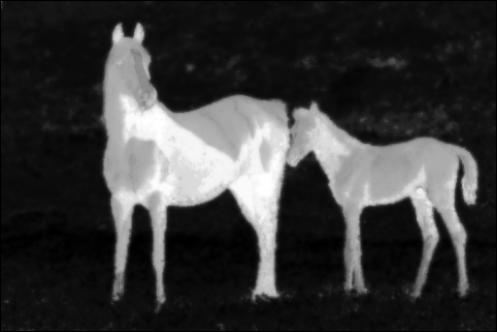}\hfill%
        \includegraphics[width=.49\linewidth]{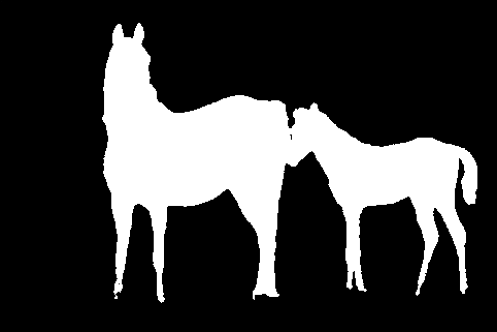}
        \\[1pt]
        \includegraphics[width=.49\linewidth]{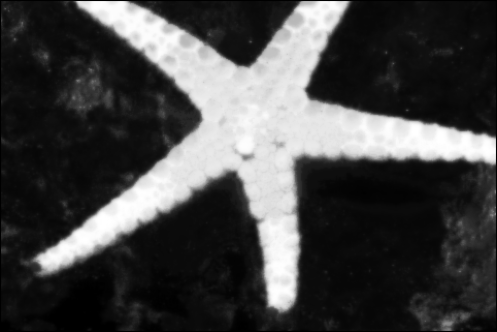}\hfill%
        \includegraphics[width=.49\linewidth]{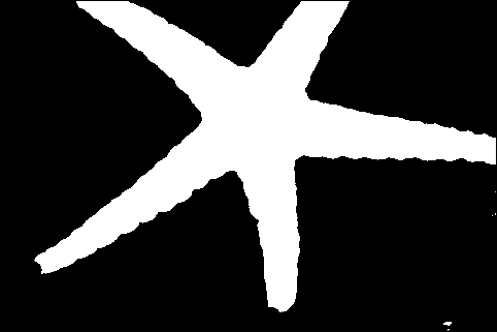}
        \caption{$\mu = 1$ ($0.17s$)}
    \end{subfigure}
    \hspace{3pt}
    \begin{subfigure}[t]{0.27\textwidth}
        \includegraphics[width=.49\linewidth]{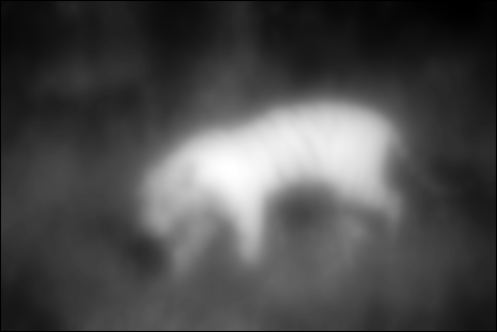}\hfill%
        \includegraphics[width=.49\linewidth]{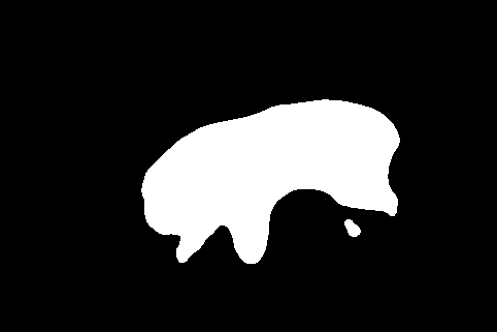}
        \\[1pt]
        \includegraphics[width=.49\linewidth]{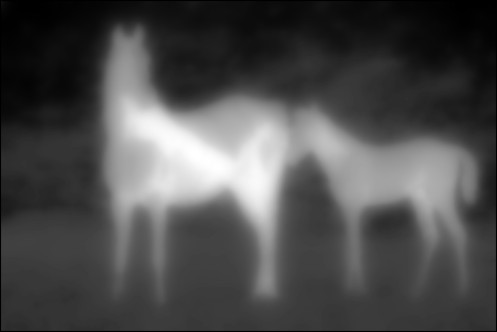}\hfill%
        \includegraphics[width=.49\linewidth]{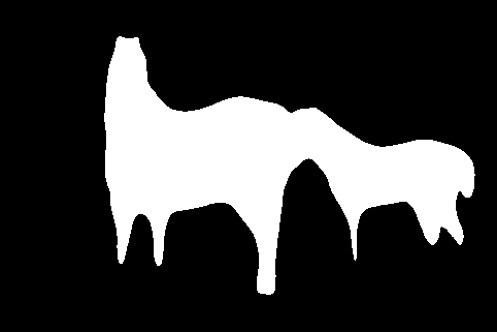}
        \\[1pt]
        \includegraphics[width=.49\linewidth]{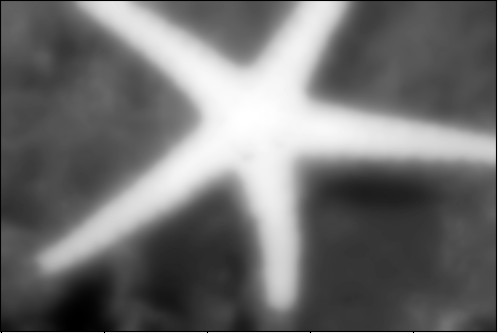}\hfill%
        \includegraphics[width=.49\linewidth]{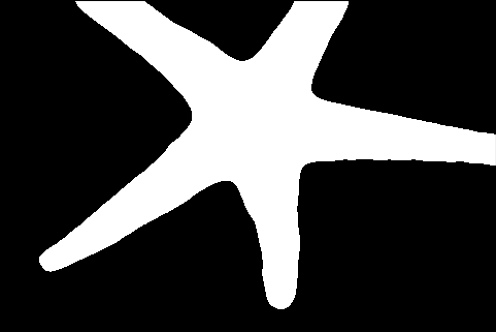}
        \caption{$\mu=50$ ($0.91s$)}
    \end{subfigure}

%% file: imgs/real/comp_real.tex
    \def\imga{1}
    \def\imgb{0} 
    \def\imgc{4}
    \def\imgd{97010}
    \def\imge{2018}
    \def\imgleg{103029}

    \begin{subfigure}[b]{0.13\textwidth}
        \includegraphics[width=\textwidth]{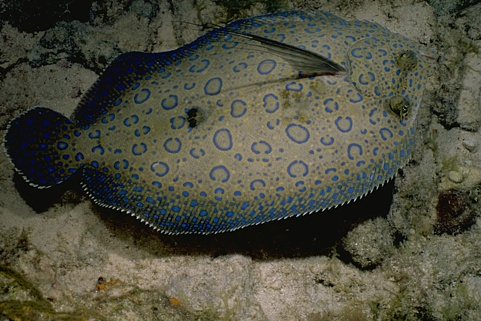} 
    \end{subfigure}%
    \hfill
    \begin{subfigure}[b]{0.13\textwidth}
        \includegraphics[width=\textwidth]{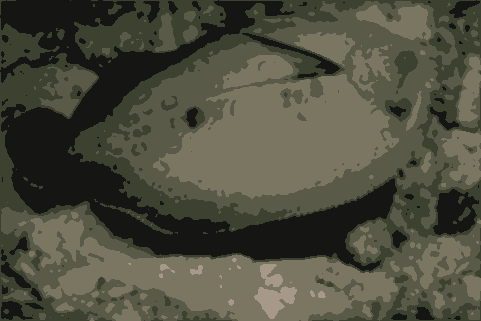}
    \end{subfigure}%
    \hfill
    \begin{subfigure}[b]{0.13\textwidth}
        \includegraphics[width=\textwidth]{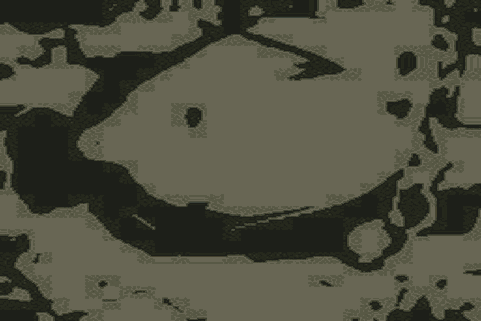}
    \end{subfigure}%
    \hfill
    \begin{subfigure}[b]{0.13\textwidth}
        \includegraphics[width=\textwidth]{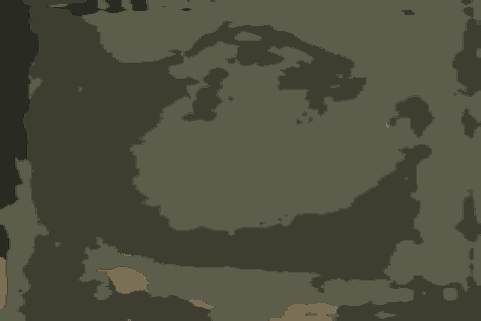}
    \end{subfigure}%
    \hfill
    \begin{subfigure}[b]{0.13\textwidth}
        \includegraphics[width=\textwidth]{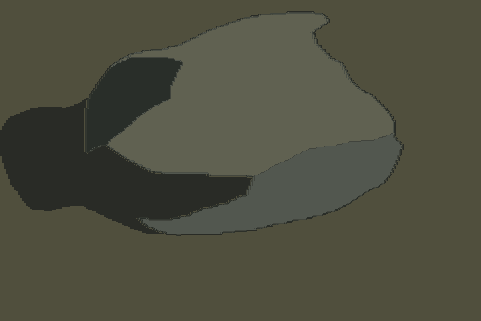}
    \end{subfigure}%
    \hfill
    \begin{subfigure}[b]{0.13\textwidth}
        \includegraphics[width=\textwidth]{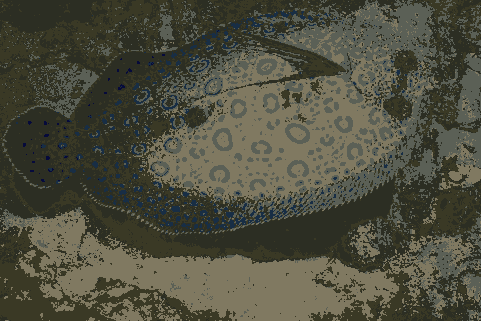}
    \end{subfigure}%
    \hfill
    \begin{subfigure}[b]{0.13\textwidth}
        \includegraphics[width=\textwidth]{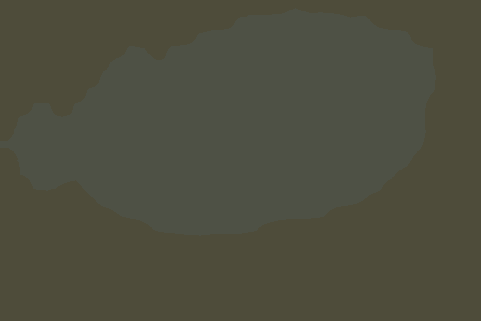}
    \end{subfigure}%

    \begin{subfigure}[b]{0.13\textwidth}
        \includegraphics[width=\textwidth]{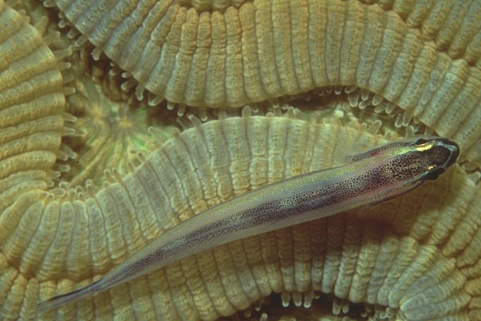} 
    \end{subfigure}%
    \hfill
    \begin{subfigure}[b]{0.13\textwidth}
        \includegraphics[width=\textwidth]{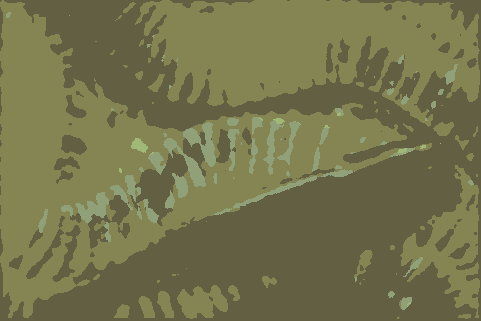}
    \end{subfigure}%
    \hfill
    \begin{subfigure}[b]{0.13\textwidth}
        \includegraphics[width=\textwidth]{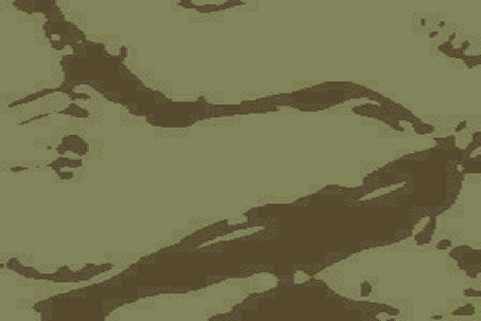}
    \end{subfigure}%
    \hfill
    \begin{subfigure}[b]{0.13\textwidth}
        \includegraphics[width=\textwidth]{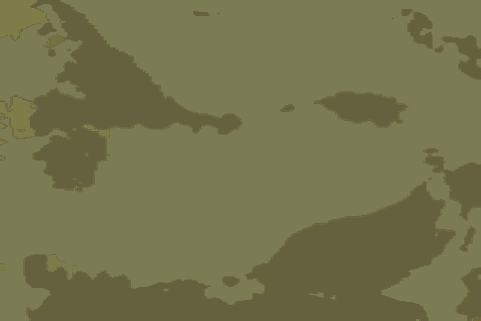}
    \end{subfigure}%
    \hfill
    \begin{subfigure}[b]{0.13\textwidth}
        \includegraphics[width=\textwidth]{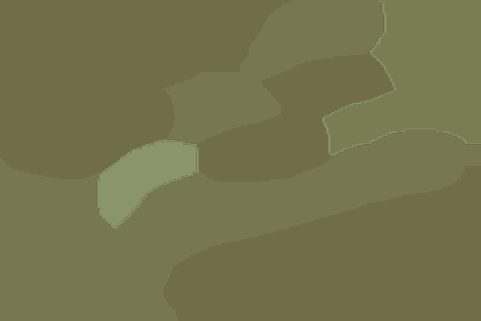}
    \end{subfigure}%
    \hfill
    \begin{subfigure}[b]{0.13\textwidth}
        \includegraphics[width=\textwidth]{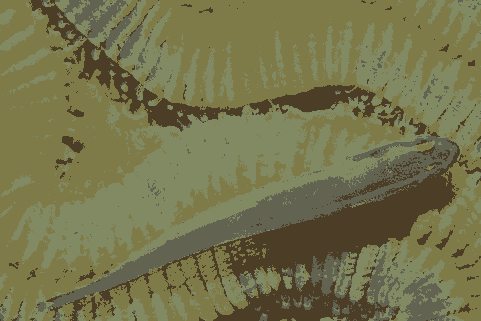}
    \end{subfigure}%
    \hfill
    \begin{subfigure}[b]{0.13\textwidth}
        \includegraphics[width=\textwidth]{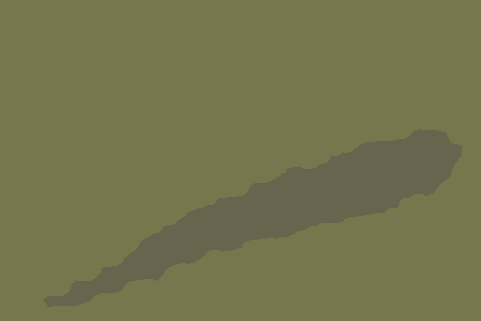}
    \end{subfigure}%

    \begin{subfigure}[b]{0.13\textwidth}
        \includegraphics[width=\textwidth]{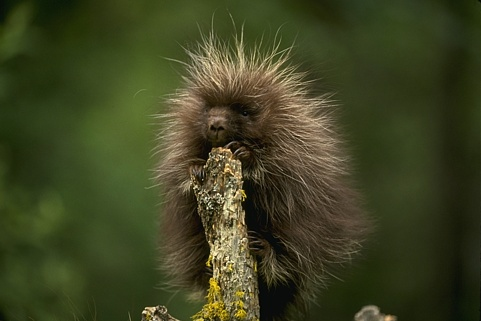} 
        \caption{Image}
    \end{subfigure}%
    \hfill
    \begin{subfigure}[b]{0.13\textwidth}
        \includegraphics[width=\textwidth]{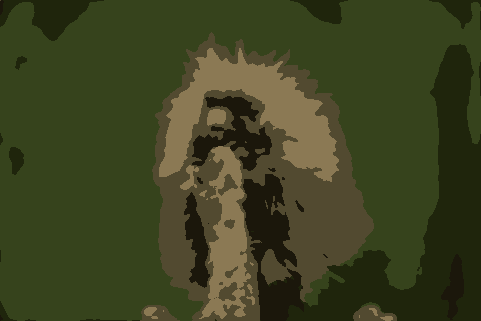}
        \caption{DFC}
    \end{subfigure}%
    \hfill
    \begin{subfigure}[b]{0.13\textwidth}
        \includegraphics[width=\textwidth]{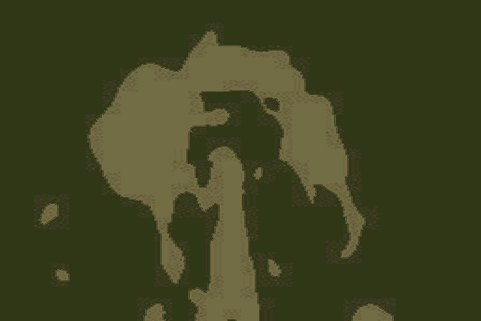}
        \caption{DoubleDIP}
    \end{subfigure}%
    \hfill
    \begin{subfigure}[b]{0.13\textwidth}
        \includegraphics[width=\textwidth]{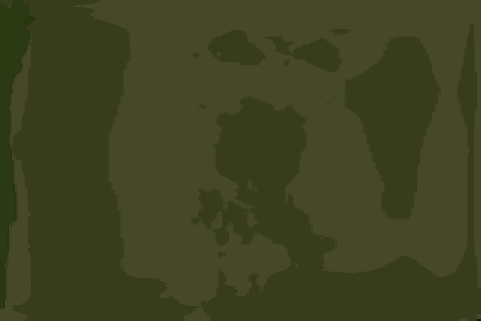}
        \caption{PiCIE}
    \end{subfigure}%
    \hfill
    \begin{subfigure}[b]{0.13\textwidth}
        \includegraphics[width=\textwidth]{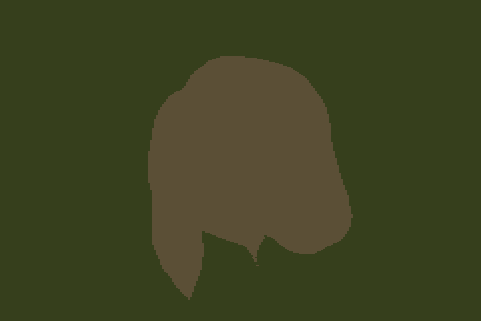}
        \caption{SegSort}
    \end{subfigure}%
    \hfill
    \begin{subfigure}[b]{0.13\textwidth}
        \includegraphics[width=\textwidth]{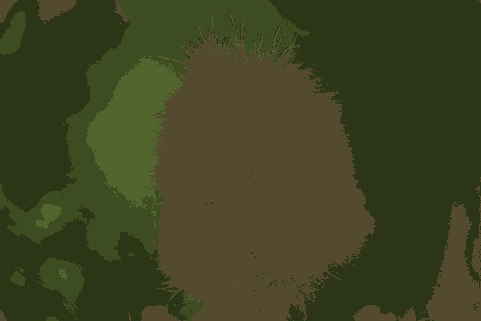}
        \caption{\textit{Ours} (RGB)}
    \end{subfigure}%
    \hfill
    \begin{subfigure}[b]{0.13\textwidth}
        \includegraphics[width=\textwidth]{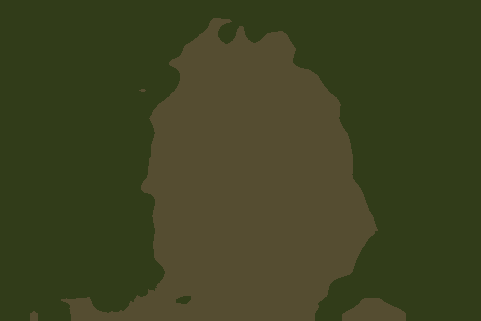}
        \caption{\textit{Ours} (DINO)}
    \end{subfigure}%